\pdfoutput=1

\documentclass[11pt]{article}

\usepackage{ACL2023}

\usepackage{times}
\usepackage{latexsym}
\usepackage{hyperref}
\usepackage{url}

\usepackage{multirow}
\usepackage{graphicx}
\usepackage{subfigure}
\usepackage{enumitem}
\usepackage{color, colortbl}
\usepackage{bm}
\usepackage{amsmath}
\usepackage{amsfonts}
\usepackage{array}
\usepackage{amsthm}
\usepackage{amssymb}
\usepackage{xspace}
\usepackage{xcolor, soul}
\usepackage{anyfontsize}
\usepackage{booktabs}
\usepackage{soul}
\sethlcolor{green}
\definecolor{OliveGreen}{rgb}{0,0.6,0}

\newcommand{\tabincell}[2]{\begin{tabular}{@{}#1@{}}#2\end{tabular}}
\newcolumntype{L}[1]{>{\raggedright\arraybackslash}p{#1}}
\newcolumntype{C}[1]{>{\centering\arraybackslash}p{#1}}
\newcolumntype{R}[1]{>{\raggedleft\arraybackslash}p{#1}}
\definecolor{Gray}{gray}{0.9}

\newcommand{\nop}[1]{}

\definecolor{mypink}{rgb}{0.858, 0.188, 0.478}

\newcommand{\model}{\textsc{BART-FG}\xspace}

\newcommand{\redbox}[2]{\colorbox[RGB]{250,206,248}{\makebox(#1,3.5){#2}}}
\newcommand{\smallredbox}[2]{\colorbox[RGB]{250,206,248}{\makebox(#1,3){#2}}}

\usepackage[T1]{fontenc}

\usepackage[utf8]{inputenc}

\usepackage{microtype}

\usepackage{inconsolata}

\title{Attacking Open-domain Question Answering by Injecting Misinformation}

\author{Liangming Pan \\
University of California, Santa Barbara \\
\texttt{liangmingpan@ucsb.edu} \And
Wenhu Chen \\
University of Waterloo \\
\texttt{wenhuchen@uwaterloo.ca} \AND
Min-Yen Kan \\
National University of Singapore \\
\texttt{kanmy@comp.nus.edu.sg} \And
William Yang Wang \\
University of California, Santa Barbara \\
\texttt{william@cs.ucsb.edu}
}

\begin{document}
\maketitle
\begin{abstract}
With a rise in false, inaccurate, and misleading information in propaganda, news, and social media, real-world Question Answering (QA) systems face the challenges of synthesizing and reasoning over \textit{misinformation-polluted contexts} to derive correct answers. This urgency gives rise to the need to make QA systems robust to misinformation, a topic previously unexplored. We study the risk of misinformation to QA models by investigating the sensitivity of open-domain QA models to corpus pollution with misinformation documents. 
We curate both human-written and model-generated false documents that we inject into the evidence corpus of QA models, and assess the impact on the performance of these systems. Experiments show that QA models are vulnerable to even small amounts of evidence contamination brought by misinformation, with large absolute performance drops on all models. Misinformation attack brings more threat when fake documents are produced at scale by neural models or the attacker targets on hacking specific questions of interest. To defend against such a threat, we discuss the necessity of building a misinformation-aware QA system that integrates question-answering and misinformation detection in a joint fashion. 
\end{abstract}

\section{Introduction}
A typical Question Answering (QA) system~\citep{DBLP:conf/acl/ChenFWB17,DBLP:conf/naacl/YangXLLTXLL19,DBLP:conf/emnlp/KarpukhinOMLWEC20,DBLP:conf/acl/YamadaAH20,DBLP:conf/naacl/GlassRCNCG22} starts by retrieving a set of relevant \textit{context documents} from the Web, which is then examined by a machine reader to identify the correct answer. Existing works typically equate Wikipedia as the web corpus. Therefore, all retrieved context documents are assumed to be clean and trustable. However, real-world QA faces a much noisier environment, where the web corpus is tainted with \textit{misinformation}. This includes unintentional factual mistakes made by human writers and deliberate disinformation intended to deceive. Aside from human-created misinformation, we are also facing the inevitability of AI-generated misinformation. With the continuing progress in text generation~\citep{radford2019language,DBLP:conf/nips/BrownMRSKDNSSAA20,DBLP:conf/acl/LewisLGGMLSZ20,DBLP:InstructGPT,DBLP:GPT4}, realistic-looking fake web documents can be generated at scale by malicious actors~\citep{DBLP:conf/nips/ZellersHRBFRC19,huang-etal-2023-faking,DBLP:RiskLLM}. 

\begin{figure*}[!t]
	\centering
	\includegraphics[width=16cm]{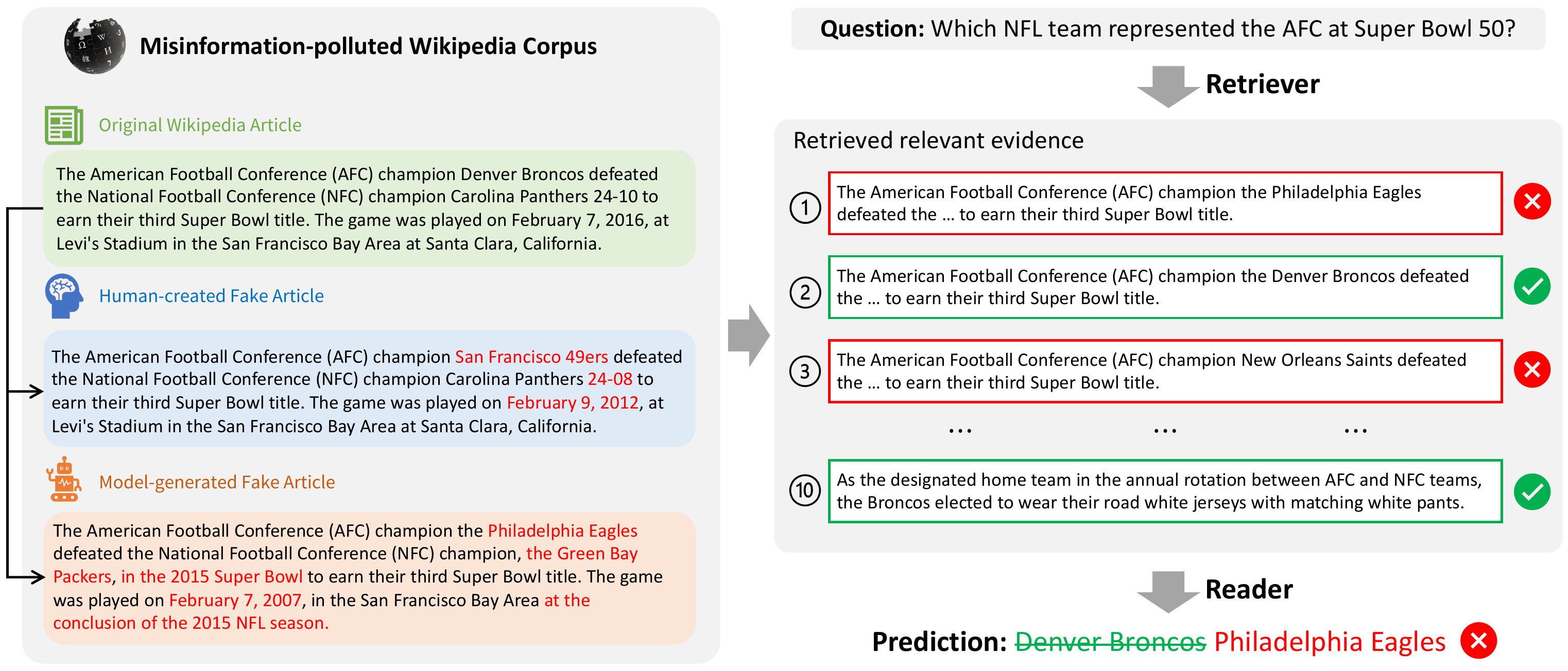}
    \caption{Our framework injects human-created and model-generated misinformation documents into the QA evidence repository (left) and evaluates the impact on the performance of open-domain QA systems (right). }
    \label{fig:general_framework}
\end{figure*}

The presence of misinformation --- no matter deliberately created or not, no matter human-written or machine-generated --- affects the reliability of the QA system by bringing in \textit{contradicting} information. As shown in Figure~\ref{fig:general_framework} (right side), when both real and fake information are retrieved as context documents, the QA models can be easily confused by the contradicting answers given by both parties, given the fact that they do not have the ability to identify fake information and reason over contradicting contexts. Although current QA models often achieve promising performance under the idealized case of clean contexts, we argue that they may easily fail under the more realistic case of misinformation-mixed contexts. 

We study the risks of misinformation to question answering by investigating how QA models behave on a \textit{misinformation-polluted web corpus} that is mixed with both real and fake information. 
To create such corpus, we propose a \textit{misinformation attack} strategy which curates fake versions of Wikipedia articles and then injects them into the clean Wikipedia corpus. For a Wikipedia article $P$, we create its fake version $P'$ by modifying information in $P$, such that: 1) certain information in $P'$ contradicts with the information in $P$, and 2) $P'$ is fluent, consistent, and looks realistic. We study both human-written and model-generated misinformation. For the human-written part, we ask Mechanical Turkers to create fake articles by modifying original wiki articles. For the model-generation part, we propose a strong rewriting model, namely \model, which can controllably mask and re-generate text spans in the original article to produce fake articles. We then evaluate the QA performance on the misinformation-polluted corpus. A robust QA model should be able to deal with misinformation and properly handle contradictory information. 

Unfortunately, from extensive experiments, we find that existing QA models are vulnerable to misinformation attacks, 
regardless of whether the fake articles are manually written or model-generated. The state-of-the-art open-domain QA pipeline, with ColBERT~\cite{DBLP:conf/naacl/SanthanamKSPZ22} as the retriever and the DeBERTa~\cite{DBLP:journals/corr/abs-2111-09543} as the reader, suffers from noticeable performance drops in five different attack modes. Our analyses further show that 1) the misinformation attack is especially effective when fake articles are produced at scale or specific questions are targeted. 2) humans do not show an obvious advantage over our \model model in creating more deceiving fake articles. 

In summary, we investigate the potential risk of open-domain QA under misinformation. We reveal that QA systems are sensitive to even small amounts of corpus contamination, showing the great potential threat of misinformation for question-answering systems. We end by discussing the necessity of building a misinformation-aware QA system. We release the data and codes publicly, helping pave the way for follow-up research in studying how to protect open-domain QA models against misinformation\footnote{\url{https://github.com/teacherpeterpan/ContraQA/}}. 

\section{Related Work}

\paragraph{Open-domain Question Answering.} To answer a question, open-domain QA systems employ a \textit{retriever-reader} paradigm that first retrieves relevant documents from a large evidence corpus and then predicts an answer conditioned on the retrieved documents. Promising advances have been made towards improving the reader models~\cite{DBLP:conf/naacl/YangXLLTXLL19,DBLP:conf/eacl/IzacardG21} and neural retrievers~\cite{DBLP:conf/acl/LeeCT19,DBLP:journals/corr/abs-2002-08909,santhanam-etal-2022-colbertv2}. However, since Wikipedia is used as the evidence corpus, previous works take for granted the assumption that the retrieved documents are trustworthy. This assumption becomes questionable with the rapid growth of fake and misleading information in the real world. In this work, we take the initiative to study the potential threat that misinformation can bring to QA systems, calling for a new direction of building misinformation-immune QA systems. 



\paragraph{Improving Robustness for QA.} 
Our work aims to analyze vulnerabilities to develop more robust QA models. Current QA models demonstrate brittleness in different aspects. QA models often rely on spurious patterns between the question and context rather than learning the desired behavior. They might ignore the question entirely~\citep{DBLP:conf/emnlp/KaushikL18}, focus primarily on the answer type~\citep{DBLP:conf/acl/MudrakartaTSD18}, or ignore the ``intended'' mode of reasoning for the task~\citep{DBLP:conf/acl/JiangB19,DBLP:conf/acl/NivenK19}. QA models also generalize badly to out-of-domain (OOD) data~\citep{DBLP:conf/acl/KamathJL20}. For example, they often make inconsistent predictions for different semantically equivalent questions~\citep{DBLP:conf/acl/GanN19,DBLP:conf/acl/RibeiroGS19}. Similar to our paper, a few prior works~\cite{chen-etal-2022-rich,DBLP:journals/corr/abs-2212-10002,DBLP:conf/uss/AbdelnabiF23} investigated the robustness of QA models under conflicting information. For example,~\citet{DBLP:journals/corr/abs-2109-05052} shows QA models are less robust to OOD data where the contextual information contradicts the learned information. Different from these works, we study from a new angle of QA robustness: the vulnerability of QA models under misinformation. 





\paragraph{Combating Neural-generated Misinformation. }
Advanced text-generation models offer a powerful tool for augmenting the training data of downstream NLP applications~\cite{DBLP:conf/acl/PanCXKW20,chen-etal-2023-empirical}. However, these models also pose a risk of being exploited for malicious activities, such as generating convincing fake news~\citep{DBLP:conf/nips/ZellersHRBFRC19}, fraudulent online reviews~\citep{DBLP:conf/emnlp/GarbaceaCYM19,DBLP:conf/aina/AdelaniMFNYE20}, and spam. Even humans find it struggle to detect such synthetically-generated misinformation~\cite{DBLP:conf/acl/ClarkASHGS20}. When produced at scale, neural-generated misinformation can pose threats to many NLP applications. For example, a recent work by~\citep{DBLP:conf/aaai/DuBM22} finds that synthetic disinformation can significantly affect the behavior of modern fact-checking systems. In this work, we study the risk of neural-generated misinformation to QA models. 


\section{Misinformation Documents Generation}


We simulate the potential vulnerability of question-answering models to corpus pollution with misinformation documents by injecting both human-written and model-generated false documents into the evidence corpus, and assess the impact on the performance of these systems. We base our study on the SQuAD 1.1~\cite{DBLP:conf/emnlp/RajpurkarZLL16} dataset, one of the most popular benchmarks for evaluating QA systems. We use all the 2,036 unique Wikipedia passages from the validation set for our study. 
For each Wikipedia passage $\mathcal{P}^R$, we create a set of $N$ fake passages $(\mathcal{P}^F_1, \cdots, \mathcal{P}^F_N)$ by modifying some information in $\mathcal{P}^R$, with the requirement that each fake passage look realistic while containing contradicting information with $\mathcal{P}^R$. 

We use two different ways to create fake passages: 1) \textbf{via human edits}: we ask online workers from Amazon Mechanical Turk (AMT) to produce fake passages by modifying the original passage, and 2) \textbf{via \model}: our novel generative model \model, which iteratively masks and re-generates text spans from the original passage to produce fake passages. 


\subsection{Manual Creation of Fake Passages}
\label{subsec:human_fake_generation}

To solicit human-written deceptive fake passages, we release 2K HITs (human intelligence tasks) on the AMT platform, where each HIT presents the crowd-worker with one passage $\mathcal{P}^R$ in the SQuAD validation set. We ask workers to modify the contents of the given passage to create a fake version, following the below guidelines:

\vspace{0.15cm}
\noindent $\bullet$ The worker should make \textit{at least $M$ edits} at different places, where $M$ equals to one plus the number of sentences in the contexts $\mathcal{C}^R$. 

\noindent $\bullet$ The worker should make at least one \textit{long edit} that rewrites at least half of a sentence. 

\noindent $\bullet$ The edits should modify key information to make it \textit{contradict with the original}, such as time, location, purpose, outcome, reason, etc. 

\noindent $\bullet$ The modified passage should be \textit{fluent and look realistic}, without commonsense errors. 
\vspace{0.15cm}

\noindent To select qualified workers, we restrict our task to workers who are located in five native English-speaking countries\footnote{Australia, Canada, Ireland, United Kingdom, USA}, and who maintain an approval rating of at least 90\%. To ensure the annotations fulfil our guidelines, we give ample examples in our annotation interface with detailed explanations to help workers understand the requirements. The detailed annotation guideline is in Appendix~\ref{sec:annotation_guideline}. We also hired three computer science major graduate students as human experts to validate a HIT's annotation. In the end, 104 workers participated in the task. The average completion time for one HIT is 5 minutes, and payment is \$1.0 U.S. dollars/HIT. The average acceptance rate was 93.75\%. 

\subsection{Model Generation of Fake Passages}

\begin{figure}[!t]
	\centering
	\includegraphics[width=7.8cm]{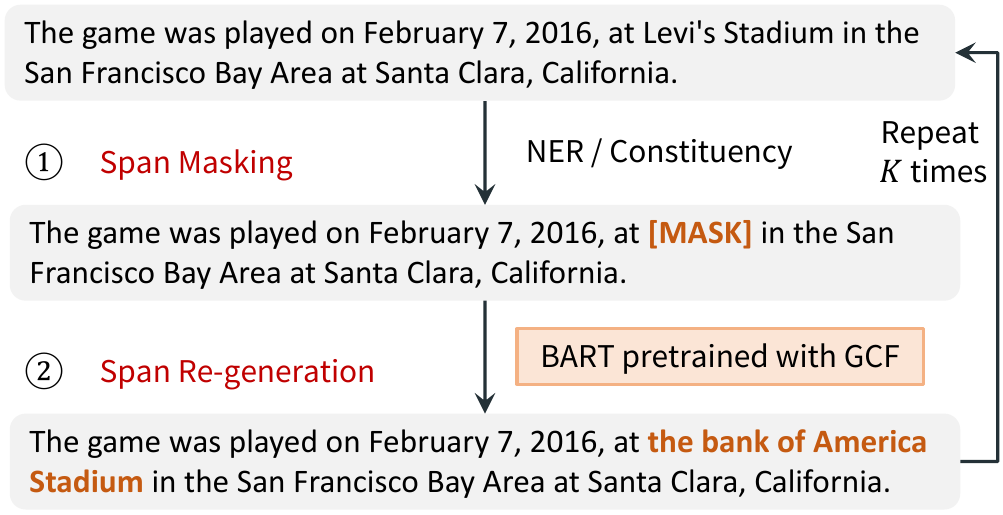}
    \caption{Overview of the \model model, illustrated by an example sentence. }
    \label{fig:bart_fg_model}
\end{figure}

Aside from human-written misinformation, we also want to explore the threat of machine-generated misinformation to QA. This source may be more of a concern than human-created misinformation, since they can easily be produced at scale. Recently introduced large-scale generative models, such as GPT2~\citep{radford2019language}, BART~\citep{DBLP:conf/acl/LewisLGGMLSZ20}, and Google T5~\citep{DBLP:journals/jmlr/RaffelSRLNMZLL20}, can produce realistic-looking texts, but they do not lend themselves to producing controllable generation that only replaces the key information with contradicting contents. Therefore, to evaluate the efficacy of realistic-looking neural fake passages, we propose \textit{BART Fake Passage Generator (\model)}, which produces both realistic and controlled generated text by iteratively modifying the original passage. As shown in Figure~\ref{fig:bart_fg_model}, for each sentence $S$ of the original passage, \model produces its fake version $S'$ via a two-step process: 


\vspace{0.15cm}

\noindent \textbf{1) Span Masking}. We first obtain a set of candidate text spans from the input sentence. We then randomly select a span and replace it with a special mask token \texttt{[MASK]}. We employ two different ways to get the candidate spans. 1) \textit{NER}: we use Spacy\footnote{\url{https://spacy.io/usage/linguistic-features\#named-entities}} to extract name entities as the candidate spans. 2) \textit{Constituency}: we apply the constituency parser implemented in AllenNLP\footnote{\url{https://demo.allennlp.org/constituency-parsing}} to extract constituency spans from the input sentence as the candidate spans. We choose to mask named entities / constituency phrases instead of random spans because: 1) they represent complete semantic units such as ``Super Bowl 50'', which avoids meaningless random phrases such as ``Bowl 50''; and 2) they often represent important information in the sentence --- such as time, location, cause, etc. 


\vspace{0.15cm}

\vspace{0.15cm}

\noindent \textbf{2) Span Re-generation}. We fill in the mask by generating a phrase different from the masked phrase. The mask is filled by the BART model fine-tuned on the Wikipedia dump with a new self-supervised task called \textit{gap span filling}, introduced later. 

\vspace{0.15cm}

The above pipeline is iteratively run for $K$ times to generate sentence $S'$ from $S$. We choose to make the edits iteratively rather than in parallel to model interaction between multiple edits. For example, in Figure~\ref{fig:bart_fg_model}, if the previous edit changes ``Santa Clara'' to ``Atlanta'', the next edit can choose to change ``California'' into ``Georgia'' to make the contents more consistent and realistic. 

\paragraph{Gap Span Filling (GSF) Pre-Training.}
To train the BART model to learn how to fill in a masked span, we propose a new pre-training task named \textit{Gap Span Filling (GSF)}. For each article in the Wikipedia dump that consists of $T$ sentences $[S_1, S_2, \cdots, S_T]$, where each sentence is a word sequence $S_t = [w^t_1, \cdots, w^t_{|S_t|}]$, we construct the following training data for $t=2, \cdots, T-1$:

\vspace{0.2cm}
{\fontsize{10.5}{10}\selectfont 

\noindent Input: $S_1, S_{t-1}, w^t_{1:a-1}, \texttt{[MASK]}, w^t_{b+1:|S_t|}, S_{t+1}$ $\quad \quad \quad$ Output: $w^t_{a:b} = [w^t_a, \cdots, w^t_b]$

}
\vspace{0.2cm}

\noindent where the output represents a masked constituency or named entity span that starts with the $a$-th word and ends with $b$-th word. The input is the concatenation of the first sentence $S_1$, the previous sentence $S_{t-1}$, the current sentence $S_t$ with one span being masked, and the subsequent sentence $S_{t+1}$. The BART model is fine-tuned to predict the output given the input on the entire Wikipedia dump. This task trains the BART model to predict the masked constituency / named entity span, given both global contexts ($S_1$) and local contexts ($S_{t-1}, S_{t+1}$). We use the \texttt{facebook/bart-large} model provided by Hugging Face (406M parameters). 

\begin{table*}[!t]
    \small
	\begin{center}
	    \renewcommand{\arraystretch}{1.1}
		\begin{tabular}{ c l l } \hline
        \textbf{\#} & \textbf{Original Contexts} & \textbf{Contradicting Contexts} \\ \hline
        (1) & \tabincell{l}{The game was played on \redbox{55}{February 7, 2016} at \redbox{50}{Levi's Stadium} \\ in the \redbox{175}{San Francisco Bay Area at Santa Clara, California.}} & \tabincell{l}{The game was played on \redbox{60}{December 7, 2015} at the \\ \redbox{165}{Bank of America Stadium in Denver, Colorado.}} \\ \hline
        (2) & \tabincell{l}{... boycotting products manufactured through child \\ labour may \redbox{75}{force these children to} turn to more \\ dangerous or strenuous professions.} & \tabincell{l}{... boycotting products manufactured through child \\ labour may \redbox{95}{prevent these children from} turn to more \\ dangerous or strenuous professions.} \\ \hline
        (3) & \tabincell{l}{Tesla worked every day from 9:00 am until 6:00 pm \\ or later.} & \tabincell{l}{Tesla worked every day \redbox{35}{but Sunday} from 9:00 am until \\ 6:00 pm or later.} \\ \hline
        (4) & \tabincell{l}{The study \redbox{155}{suggests that boycotts are ``blunt instruments} \\ 
        \redbox{185}{with long-term consequences, that can actually harm} \\ 
        \redbox{140}{rather than help the children involved.''}} & \tabincell{l}{The study \redbox{160}{did not find any major negative repercussions} \\ \redbox{190}{from boycotts, however, and found that boycotting is} \\ \redbox{60}{the best solution.}} \\ \hline
        (5) & \tabincell{l}{A key distinction between analysis of algorithms and \\ complexity theory is that the \smallredbox{20}{former} is devoted to ..., \\ whereas the \smallredbox{15}{later} asks a more general question of ...} & \tabincell{l}{A key distinction between analysis of algorithms and \\ complexity theory is that the \smallredbox{15}{later} is devoted to ..., \\ whereas the \smallredbox{20}{former} asks a more general question of ...} \\ \hline
        (6) & \tabincell{l}{On the whole, Eisenhower's support of the nation's \\ fledgling \redbox{48}{space program} was officially \redbox{20}{modest} until the \\ \redbox{115}{Soviet launch of Sputnik in 1957}, gaining the Cold War 
        \\ enemy enormous prestige \redbox{60}{around the world.} } & \tabincell{l}{On the whole, Eisenhower's support of the nation's \\ fledgling \redbox{30}{MK Ultra} was officially \redbox{35}{terminated} until the \\ \redbox{70}{Cuban missile crisis}, gaining the Cold War enemy \\ enormous admiration in \redbox{80}{less developed nations.}} \\ \hline
		\end{tabular}
	\end{center}
\caption{Examples of original passages and their corresponding fake versions, where the information changes are highlighted. These examples represent six common types of created misinformation. }
\label{tbl:dataset_examples}
\end{table*}

\subsection{Analysis of the Generated Fake Passages}


Table~\ref{tbl:dataset_examples} shows examples from six original passages with their corresponding fake versions, which represent six common types of modifications made by the human and the model, explained as follows: 

\vspace{0.1cm}

\noindent (1) \textbf{Entity Replacement}: replacing entities (\textit{e.g.}, person, location, time, number) with other entities with the same type, a common type of modification for both human edits and \model. 

\noindent (2) \textbf{Verb Replacement}: replacing verb or verb phrase with its antonymic  meaning, \textit{e.g.}, ``force these children to'' $\rightarrow$ ``prevent these children from''.  

\noindent (3) \textbf{Adding Restrictions}: create contradiction by inserting additional restrictions to the original content, \textit{e.g.}, ``every day'' $\rightarrow$ ``every day but Sunday''. 

\noindent (4) \textbf{Sentence Rephrasing}: rewrite the whole sentence to express a contradicting meaning, exemplified by (4). This is common in human edits but rarely seen in model-generated passages, since this requires deep reading comprehension.   

\noindent (5) \textbf{Disrupting Orders}: make a contradiction by disrupting some property of the entities;
\textit{e.g.}, example (5) switches the property of ``analysis of algorithms'' and ``complexity theory''. 

\noindent (6) \textbf{Consecutive Replacements}: humans are better in making consecutive edits to create a contradicting yet coherent sentences, exemplified by (6). 

\section{Corpus Pollution with Misinformation}

Given the fake passages curated by both human and our \model model, we now study how extractive QA models behave under an evidence corpus that is polluted with misinformation. We begin with creating a \textit{\textbf{clean corpus}} for question answering which contains one million real Wikipedia passages. We obtain the Wikipedia passages from the 2019/08/01 Wikipedia dump provided by the Knowledge-Intensive Language Tasks (KILT) benchmark~\cite{DBLP:conf/naacl/PetroniPFLYCTJK21}, in which the Wikipedia articles have been pre-processed and separated into paragraphs. We sample 1M paragraphs from KILT and ensure that all the 20,958 Wikipedia passages in the SQuAD dataset are included in the corpus. We then explore the following five ways of polluting the clean corpus with human-created and synthetically-generated false documents. 

\vspace{0.15cm}

\noindent $\bullet$ \textbf{Polluted-\textit{Human}}. In Section~\ref{subsec:human_fake_generation}, we asked human annotators to create a fake version for each passage in the SQuAD dev set. We inject those 2,023 fake passages into the clean corpus. 

\noindent $\bullet$ \textbf{Polluted-\textit{NER}}. We use \model to generate 10 fake passages for each real passage in the SQuAD dev set, using NER to get candidate spans. We mask and re-generate all candidate spans to create each fake passage. Nucleus sampling~\cite{DBLP:conf/iclr/HoltzmanBDFC20} is used to ensure diversity in generation, giving us 18,233 non-repetitive fake passages in total. We inject them into the clean corpus. 

\noindent $\bullet$ \textbf{Polluted-\textit{Constituency}}. We generate 10 fake passages for each real passage using constituency parsing to get candidate spans in \model. Since there are far more constituency phrases than named entities in a sentence, to ensure efficiency, we fix the number of replacements $K=3$ for each sentence. We get 19,796 non-repetitive fake passages and inject them into the clean corpus. 

\noindent $\bullet$ \textbf{Polluted-\textit{Hybrid}}. We inject all of the above-generated fake passages into the clean corpus. 

\noindent $\bullet$ \textbf{Polluted-\textit{Targeted}}. 
In the above settings, the attacker (human or \model model) tries to create misleading fake information \textit{without} knowing the target questions. However, in another attack mode, attackers have \textit{particular questions of interest} that they want to mislead the QA system into getting wrong answers. 
To explore how QA systems react to such attacks, in this setting we assume the attacker targets the questions in the SQuAD dev set. We then create fake passages by masking and re-generating the \textit{answer spans} of these questions using \model. Through this, we get 10,101 fake passages and insert them into the clean corpus. 


\begin{table*}[!t]
\centering
\resizebox{\textwidth}{!}{
\renewcommand{\arraystretch}{1.1}
\begin{tabular}{l|C{1.1cm}C{1.1cm}|C{1.1cm}C{1.1cm}|C{1.1cm}C{1.1cm}|C{1.1cm}C{1.1cm}|C{1.1cm}C{1.1cm}}
\toprule
  \multirow{2}{*}{\textbf{Evidence Corpus}} & 
  \multicolumn{2}{c|}{\tabincell{c}{\textbf{RoBERTa} \\ \small \citep{DBLP:journals/corr/abs-1907-11692}}} &
  \multicolumn{2}{c|}{\tabincell{c}{\textbf{SpanBERT} \\ \small \citep{DBLP:journals/tacl/JoshiCLWZL20}}} & \multicolumn{2}{c|}{\tabincell{c}{\textbf{Longformer} \\ \small \citep{DBLP:journals/corr/abs-2004-05150}}} &
  \multicolumn{2}{c|}{\tabincell{c}{\textbf{ELECTRA} \\ \small \citep{DBLP:conf/iclr/ClarkLLM20}}} &
  \multicolumn{2}{c}{\tabincell{c}{\textbf{DeBERTaV3} \\ \small \citep{DBLP:journals/corr/abs-2111-09543}}} \\ \cmidrule(lr){2-3} \cmidrule(lr){4-5} \cmidrule(lr){6-7} \cmidrule(lr){8-9} \cmidrule(lr){10-11}
  & \textbf{EM} & \textbf{F1} & \textbf{EM} & \textbf{F1} & \textbf{EM} & \textbf{F1} & \textbf{EM} & \textbf{F1} & \textbf{EM} & \textbf{F1} \\ \midrule
Clean & 53.72 & 59.45 & 55.58 & 61.30 & 56.40 & 61.68 & 55.41 & 61.52 & 62.30 & 67.85 \\ \midrule
Polluted-\textit{Human} & 48.47 & 56.84 & 51.20 & 58.26 & 52.39 & 59.03 & 51.43 & 59.04 & 58.16 & 64.82 \\ 
Polluted-\textit{Constituency} & 46.07 & 54.63 & 46.47 & 55.38 & 47.69 & 56.07 & 45.84 & 55.05 & 50.88 & 59.63 \\ 
Polluted-\textit{NER} & 42.23 & 50.34 & 44.01 & 52.64 & 45.25 & 53.50 & 43.40 & 52.54 & 48.74 & 57.16 \\ 
Polluted-\textit{Hybrid} & 41.96 & 50.17 & 44.18 & 53.61 & 44.93 & 53.98 & 42.69 & 52.81 & 48.14 & 57.63 \\ \midrule
Polluted-\textit{Targeted} & 25.29 & 34.22 & 25.55 & 34.76 & 26.92 & 35.84 & 25.42 & 34.80 & 29.52 & 38.80 \\ \bottomrule
\end{tabular}%
}
\caption{Effects of different modes of misinformation attacks on the open-domain QA performance in SQuAD.}  
\label{tbl:contradict_qa_results}
\end{table*}

\section{Models and Experiments}

We now how question answering models behave under such misinformation-polluted environment. To answer a given question, the QA systems employ a \textit{retrieve-then-read} pipeline that first retrieves $N$ (we set $N=5$) relevant contextual documents from the evidence corpus and then predicts an answer conditioned on the retrieved documents. For document retrieval, we apply the widely-used sparse retrieval based on BM25, implemented with the Pyserini toolkit~\cite{DBLP:conf/sigir/LinMLYPN21}. For question answering, we consider five state-of-the-art QA models with public code that achieved strong results on the public leader board of SQuAD: \textit{RoBERTa-large}~\citep{DBLP:journals/corr/abs-1907-11692}, \textit{Span-BERT}~\citep{DBLP:journals/tacl/JoshiCLWZL20}, \textit{Longformer}~\citep{DBLP:journals/corr/abs-2004-05150}, \textit{ELECTRA}~\citep{DBLP:conf/iclr/ClarkLLM20}, and \textit{DeBERTa-V3}~\citep{DBLP:journals/corr/abs-2111-09543}. We use their model checkpoints fine-tuned on the SQuAD training set from the Hugging Face library. We use the standard Exact Match (EM) and $F_1$ metrics to measure QA performance. 

\subsection{Main Results}
\label{subsec:main_results}

In Table~\ref{tbl:contradict_qa_results}, we show the performance of different QA models on the SQuAD dev set under the clean evidence corpus (\textit{Clean}) and the performance under the misinformation-polluted corpus (\textit{Polluted}). We have two major observations. 


For all models, we see a noticeable performance drop when generated fake passages are introduced into the clean evidence corpus: the smallest average performance drop is 7.72\% in relative EM value (Polluted-\textit{Human}), while the largest drop is 53.19\% (Polluted-\textit{Targeted}). This indicates that \ul{QA models are sensitive to misinformation attack}; even limited amounts of injected fake passages comprising 0.2\% (\textit{Human}) to 4.0\% (\textit{Hybrid}) of the entire corpus can noticeably affect downstream QA performance. 
It reveals the potential threat of misinformation to current QA systems, given the fact that they are not trained to differentiate misinformation. 


\begin{table*}[!t]
\centering
\resizebox{\textwidth}{!}{
\renewcommand{\arraystretch}{1.1}
\begin{tabular}{l|cccccc|cccccc}
\toprule
  & \multicolumn{6}{c|}{\textbf{BM25 + DeBERTa-V3}} & \multicolumn{6}{c}{\textbf{ColBERT-V2 + DeBERTa-V3}} \\ \cmidrule{2-7} \cmidrule{8-13}
  \textbf{Evidence Corpus} & \textbf{R@1} & \textbf{R@5} & \textbf{F@1} & \textbf{F@5} & \textbf{EM} & \textbf{F1} & \textbf{R@1} & \textbf{R@5} & \textbf{F@1} & \textbf{F@5} & \textbf{EM} & \textbf{F1} \\ \midrule
Clean & 57.46 & 75.97 & --- & --- & 62.30 & 67.85 & 59.30 & 80.40 & --- & --- & 67.54 & 73.17 \\ \midrule
Polluted-\textit{Human} & 47.24 & 74.21 & 7.11 & 44.58 & 58.16 & 64.82 & 41.95 & 75.91 & 11.07 & 43.71 & 59.02 & 65.23 \\ 
Polluted-\textit{Constituency} & 30.21 & 49.50 & 23.64 & 46.54 & 50.88 & 59.63 & 28.63 & 47.50 & 25.01 & 48.00 & 49.17 & 58.66 \\ 
Polluted-\textit{NER} & 28.30 & 48.88 & 21.33 & 48.79 & 48.74 & 57.16 & 25.88 & 44.34 & 22.86 & 50.01 & 46.41 & 54.31 \\ 
Polluted-\textit{Hybrid} & 25.67 & 45.60 & 26.53 & 53.45 & 48.14 & 57.63 & 23.01 & 42.69 & 23.80 & 55.12 & 45.46 & 54.03 \\ \midrule
Polluted-\textit{Targeted} & 15.04 & 45.70 & 46.60 & 72.86 & 29.52 & 38.80 & 16.90 & 40.09 & 47.27 & 74.56 & 28.93 & 37.12 \\ \bottomrule
\end{tabular}%
}
\caption{Effects of different modes of misinformation attacks on the \textit{BM25} and \textit{ColBERT-V}2 retrievers.}
\label{tbl:retrieval_results}
\end{table*}

Polluted-\textit{Targeted} causes a more significant performance drop compared to the most effective question-agnostic attack (Polluted-\textit{Hybrid}) ($\sim$53\% v.s. $\sim$22\% relative EM drop), indicating that \ul{QA models are more vulnerable under question-targeted misinformation attack.} This reveals that the misinformation attack brings more threat when the attacker wants to alter the answers produced by QA systems for particular questions of interest. 
For the other four question-agnostic settings where the pollution is not targeted on specific questions, we still observe a noticeable EM drop ($\sim$20\%) for all models. Among them, Polluted-\textit{NER} causes more performance drop than Polluted-\textit{Constituency}, showing that generating misinformation by replacing named entities is more effective than replacing constituency spans. This is probably due to the nature of the SQuAD dataset, where most of the answer spans are named entities. 


\subsection{Impact of misinformation on retriever}

The success of the misinformation attack relies on the premise that fake passages can be retrieved from the polluted corpus by the retriever. To validate this, we first define a fake passage $P$ as the \textit{misleading evidence} for the question $Q$ if $P$ contains a fabricated answer for $Q$. We then report in Table~\ref{tbl:retrieval_results} the percentage of misleading evidence in the top-$k$ retrieved passages (F@k, for $k \in \{1, 5\}$) for the BM25 retriever. We find that both F@1 and F@5 are very high, while the likelihood of the ground-truth true evidence appearing in the top-1 (R@1) and top-5 (R@5) decreases significantly for polluted corpus. The results show that \ul{the injected fake passages can be easily retrieved as evidence for downstream question answering.} QA models, without the fact-checking capability, can thus be easily misled by such misinformation. 

However, BM25 only relies on syntactic features and cannot be optimized for specific tasks. Is the misinformation attack also effective for trainable dense retrievers? To explore this, we use ColBERT-V2~\cite{DBLP:conf/naacl/SanthanamKSPZ22}, the state-of-the-art dense retriever that independently encodes the question and the passage using BERT and then employs a late interaction architecture to model their similarity. 
We use the ColBERT pretrained on MSMARCO~\cite{DBLP:conf/nips/NguyenRSGTMD16} and fine-tune it with (question, context) pairs from SQuAD training set as positive samples and (question, random context) as negative samples. The retrieval and QA performance are reported in Table~\ref{tbl:retrieval_results}. 

We find that misinformation attack also affects the ColBERT retriever, decreasing R@1 and R@5 for all settings, with high percentage of fake passages being retrieved as reflected by F@1 and F@5. The results also suggest that \ul{ColBERT is less resistant to misinformation attack compared to BM25.} In the clean corpus, ColBERT outperforms BM25 in both the retrieval and the downstream QA performance. However, in all polluted corpus, the relative performance drop for ColBERT is larger than the drop for BM25. The possible explanation is: without the ability to identify fake information, a more ``accurate'' retriever tends to retrieve more seemingly relevant but false documents, making it less robust to misinformation attack. 


\begin{figure}[!t]
	\centering
	\includegraphics[width=7.5cm]{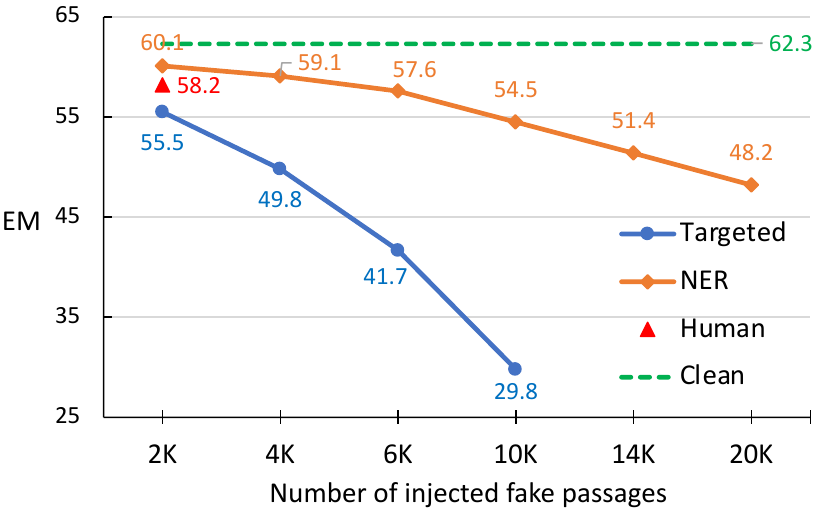}
    \caption{The EM score for DeBERTa-V3 model with different number of injected fake passages $N$. } 
    \label{fig:impact_of_fake_number}
\end{figure}


\subsection{Impact of the size of injected fake passages}

As confirmation that misinformation attacks work as expected, we depict in Figure~\ref{fig:impact_of_fake_number} how the DeBERTa model performance changes when different number of fake passages are injected into the evidence corpus. We find that the EM score steadily drops with more fake passages for both the question-targeted attack (\textit{Targeted}) and the question-agnostic attack (\textit{NER}). However, the former causes a much sharper trend of decrease, which further validates that misinformation attack is more deadly with a better knowledge of the target questions. 
Through this study, we conclude that \ul{misinformation may have a more severe impact on QA systems when they are produced at scale}. With the availability of pretrained text generation models, producing fluent and realistic-looking contexts now has a little marginal cost. This brings an urgent need to effectively defend against neural-generated misinformation in question answering. 


\subsection{Which is more deceiving: human- or model-generated misinformation?}
\label{subsec:which_more_deceiving}

We then investigate which is more deceiving to QA models: human or neural misinformation? 
To study this, we let the QA model to answer each question $Q$ under the context $\mathcal{C} = \{ \mathcal{P^R, P^H, P^C, P^N} \}$, where $P^R$ is the real passage that contains the correct answer, and $\mathcal{P^H, P^C, P^N}$ are the corresponding fake versions of $P^R$ produced by human, \model (NER), and \model (Constituency), respectively. We then analyze the source (which fake passage) of the incorrect answer when the model makes an error. If all three methods create equally deceiving fake passages, we expect to observe a uniform distribution of the error sources. 

\begin{figure}[!t]
	\centering
	\includegraphics[width=6.5cm]{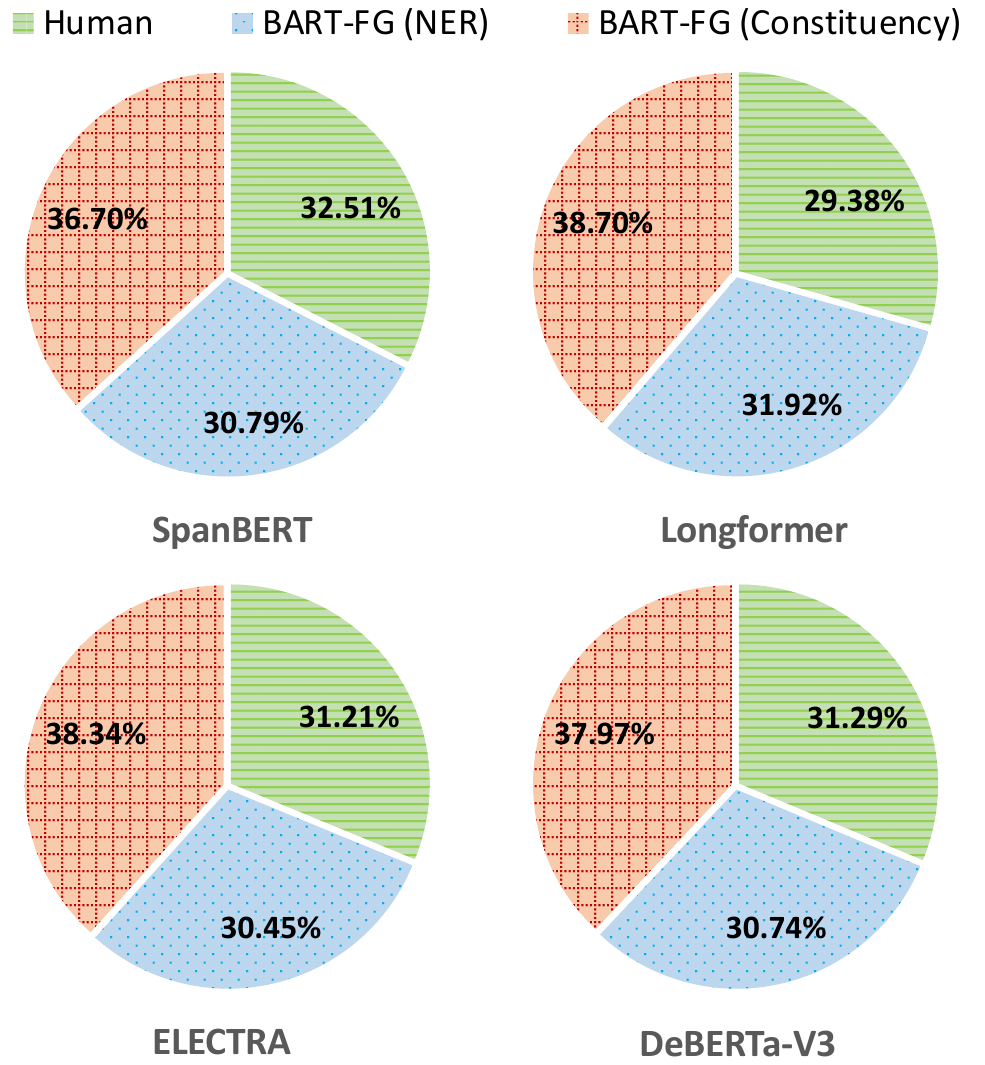}
	\caption{Distribution of error sources when the model is misled by a fake passage and gives a wrong answer. }
	\label{fig:distractor_distribution}
\end{figure}

The distribution of error sources in Figure~\ref{fig:distractor_distribution} shows that the most wrong answers are extracted from the model-generated fake passage. 
\ul{Human-created fake passages do not show an advantage over BART-FG in deceiving the QA models.} This is counter-intuitive to what we find in Table~\ref{tbl:dataset_examples} that humans make more subtle edits that require a deep level of reading comprehension, such as switching ``former'' and ``latter'' (Example~4), and changing ``every day'' to ``every day but Sunday'' (Example~3). A possible reason is that most questions in SQuAD are shallow in reasoning~\cite{DBLP:conf/acl/DuSC17}. Therefore, replacing named entities/constituency phrases is sufficient in misleading QA models into getting the wrong answers for those questions. 


\subsection{Can misinformation deceive humans?}

After showing the impact of misinformation attacks on QA systems, one natural question would be whether humans can also be distracted by misinformation during QA. To investigate this, we ran a study on Mechanical Turk where we presented crowd-workers with 500 randomly-sampled (question, context) pairs from the data in Section~\ref{subsec:which_more_deceiving}, \textit{i.e.}, each context consists of the real passage along with three fake passages created by different methods. We call this test set \textit{MisinfoQA-noisy} and the workers are asked to answer each of its questions. For comparison, we create another test set \textit{MisinfoQA-clean} where each real passage is paired with three randomly sampled other Wikipedia passages. 

\begin{table}[!t]
\centering
\resizebox{0.48\textwidth}{!}{
\renewcommand{\arraystretch}{1.1}
\begin{tabular}{l|cc|cc}
\toprule
  \multirow{2}{*}{\textbf{Setting}} & \multicolumn{2}{c|}{\textbf{MisinfoQA-\textit{noisy}}} & \multicolumn{2}{c}{\textbf{MisinfoQA-\textit{clean}}} \\ \cmidrule{2-5}
 & \textbf{EM} & \textbf{F1} & \textbf{EM} & \textbf{F1} \\ \midrule
Human & 69.13 & 78.25 & 86.57 & 91.40\\ \midrule
RoBERTa & 61.20 & 70.44 & 77.06 & 83.88 \\ 
SpanBERT & 64.00 & 72.32 & 81.65 & 88.55 \\ 
Longformer & 67.83 & 75.15 & 82.80 & 90.72 \\ 
ELECTRA & 64.21 & 72.90 & 78.27 & 86.49 \\ 
DeBERTa-V3 & 75.00 & 82.70 & 87.25 & 92.90 \\ \bottomrule
\end{tabular}%
}
\caption{QA performance under the reading comprehension settings with \textit{clean} and \textit{noisy} contexts.}
\label{tbl:human_results}
\vspace{-0.5cm}
\end{table}

Table~\ref{tbl:human_results} reports the EM and F1 for both human and different QA models. We find that all QA models suffer a large performance drop ($\sim$20\% in EM) in \textit{MisinfoQA-noisy} compared to \textit{MisinfoQA-clean}, showing that the models are largely distracted by the fake contexts rather than by the presence of additional contexts. Humans obtained an EM of 69.13 in \textit{MisinfoQA-noisy}, which, though higher than most QA models' performance, also shows a significant drop when compared to the \textit{MisinfoQA-clean} setting (86.57 EM). This shows that \ul{humans are also likely distracted by misinformation in QA}, which demonstrates the challenge of distinguishing misinformation in question answering for lay readers, the quality of the generated fake passages, and the difficulty of detecting such an attack. 




\section{Discussion and Future Work}

Finally, we discuss three possible ways to defend the threat of misinformation for QA. 

\vspace{0.1cm}

\noindent \textbf{Knowledge source engineering.} Despite being a trustful knowledge source, Wikipedia is insufficient to fulfill all the information needed in real-life question answering. Therefore, recent works~\cite{DBLP:journals/corr/abs-2112-09924} started to use the web as the QA corpus. However, when transitioning to a web corpus, we no longer have the certainty that any document is truthful. 
Therefore, the corpora will require more careful curation to avoid misinformation. This also brings the need for future retrieval models to have the ability to assess the quality of the retrieved documents and prioritize more trustworthy sources. 

\vspace{0.1cm}

\noindent \textbf{Integrating fact-checking and QA.} 
With the rise of misinformation online, automated fact-checking has received growing attention in NLP~\cite{DBLP:journals/tacl/GuoSV22}. Integrating fact-checking models into the pipeline of open-domain QA could be an effective countermeasure to misinformation, a direction neglected by prior works. A possible way is to detect potential false claims in retrieved contexts and lower their importance in downstream QA models. 

\vspace{0.1cm}

\noindent \textbf{Reasoning under contradicting contexts.}
It is common for humans to deal with contradictory information during information search. With the presence of inaccurate and false information online, future models should focus on the ability to synthesize and reason over contradicting information to derive correct answers. 

\section{Conclusion}


In this work, we evaluate the robustness of open-domain question-answering models when we contaminate the evidence corpus with misinformation. We studied two representative sources of misinformation: human-written disinformation and the misinformation-generated NLG models. Our studies reveal that QA models are indeed vulnerable under misinformation-polluted contexts. We also show that our \model model can produce fake documents at scale that are as deceptive as humans. This poses a threat to current open-domain QA models in defending neural misinformation attacks. 




\section*{Limitations}
We identify two main limitations to our study. First, although SQuAD is a typical dataset for evaluating open-domain QA models, most of the SQuAD questions are factoid and shallow in reasoning, making it relatively easy to generate misinformation targeted at SQuAD. Our results show that \model with named entity replacement can generate fake passages as deceptive as humans. However, the impact of model-generated misinformation may be over-estimated on the shallow factoid questions in SQuAD. Therefore, more QA datasets should be considered in future works, especially non-factoid questions with deeper reasoning. 

Second, this work creates misinformation by revising key information of real articles in Wikipedia. However, there are other types of misinformation in the real world, such as hoaxes, rumors, or false propaganda. However, our proposed attack model can be easily generalized to study the threat of misinformation in other domains and in other forms. 


\section*{Ethics Statement}

We plan to publicly release the human- and model-generate fake documents and open-source the code and model weights for our \model model. We note that open-sourcing the \model model may bring the potential for deliberate misuse to generate disinformation for harmful applications. The human-written and model-generated fake documents can also be misused to generate disinformation. We deliberated carefully on the reasoning for open-sourcing and share here our three reasons for publicly releasing our work. 

First, the danger of \model in generating disinformation is limited. Disinformation is a subset of misinformation that is spread deliberately to deceive. Although we utilize the innate ``hallucination'' ability of current pretrained language models to create misinformation, our model are not specialized to generate harmful disinformation such as hoaxes, rumors, or false propaganda. Instead, our model focuses on generating conflicting information by iteratively editing the original passage to test the robustness of QA to misinformation. 

Second, our model is based on the open-sourced BART model, which makes our model easy to replicate even without the released code. Given the fact that our model is a revised version of an existing publicly available model, it is unnecessary to conceal code or model weights. 

Third, our decision to release follows the similar stance of the full release of another strong detector and state-of-the-art generator of neural fake news: Grover~\citep{DBLP:conf/nips/ZellersHRBFRC19}\footnote{\url{https://thegradient.pub/why-we-released-grover/}}. The authors claim that to defend against potential threats, we need threat modeling, in which a crucial component is a strong generator or simulator of the threat. In our work, we build an effective threat model for QA under misinformation. Followup research can build on our model transparency, further enhancing the threat model. 

\section*{Acknowledgements}
This work was supported by the National Science Foundation Award \#2048122. The views expressed are those of the authors and do not reflect the official policy or position of the US government. 

\bibliography{anthology,custom}
\bibliographystyle{acl_natbib}

\clearpage

\appendix



\section{Human Annotation Guideline}
\label{sec:annotation_guideline}

\subsection{Job Description}

Given a paragraph from Wikipedia, modify some information in the paragraph to create a fake version of it. Here are the general requirements: 

\noindent $\bullet$ You should make \textit{at least $M$ edits} at different places, where $M$ is determined by the length of the passage and will show on the screen when you annotate each passage. 

\noindent $\bullet$ You should make at least one \textit{long edit} that rewrites at least half of a sentence. 

\noindent $\bullet$ The edits should modify key information to make it \textit{contradict with the original}, such as time, location, purpose, outcome, reason, etc. 

\noindent $\bullet$ The modified paragraph should be \textit{fluent and look realistic}, without commonsense errors. 

\subsection{Detailed Requirements}

\begin{figure*}
    \centering
    \includegraphics[width=14cm]{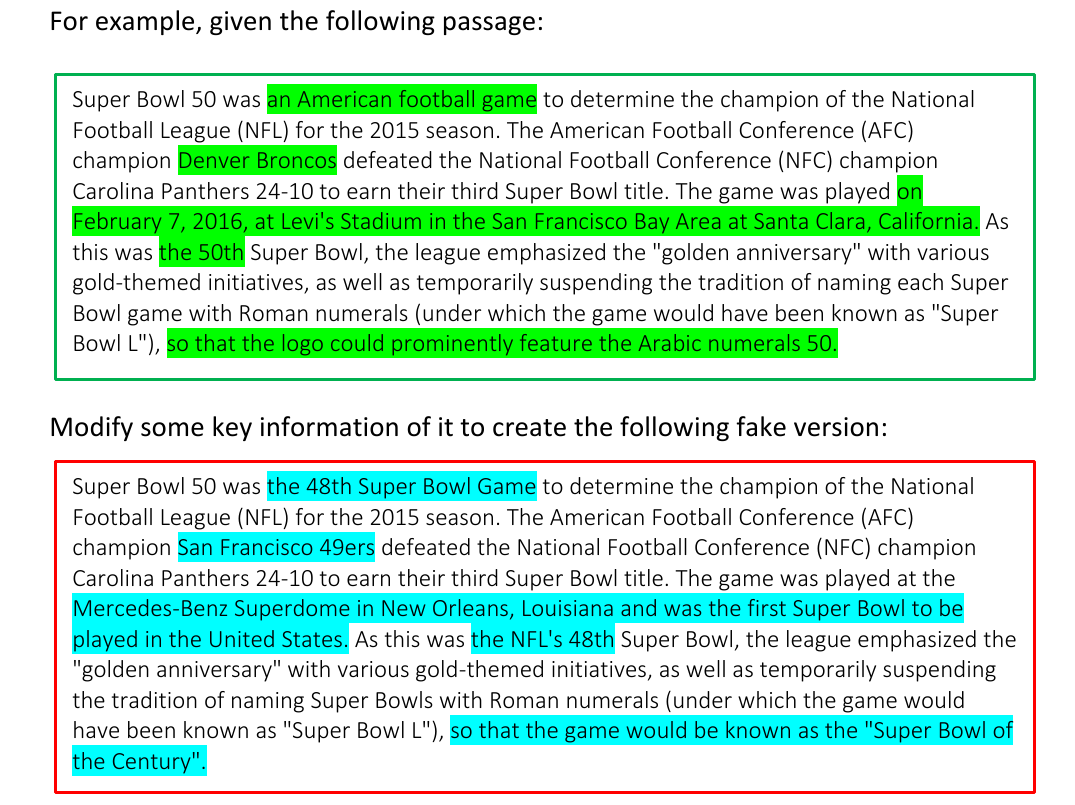}
    \caption{An example of human annotation that follows all instructions of the annotation guideline. }
    \label{fig:annotation_example}
\end{figure*}

Figure~\ref{fig:annotation_example} shows an example of modifications that fulfill all the annotation requirements. Detailed annotation instructions are as follows. 

\paragraph{1) At least make N edits at different places.}

In the above example, there are a total of 5 edits: 
\begin{itemize}
    \item ``an American football game'' $\rightarrow$ ``the 48th Super Bowl Game''
    \item ``Denver Broncos'' $\rightarrow$ ``San Francisco 49ers''
    \item ``on February 7, 2016, at Levi's Stadium in the San Francisco Bay Area at Santa Clara, California.'' $\rightarrow$ ``Mercedes-Benz Superdome in New Orleans, Louisiana and was the first Super Bowl to be played in the United States.''
    \item ``the 50th'' $\rightarrow$ ``the NFL’s 48th''
    \item ``so that the logo could prominently feature the Arabic numerals 50.'' $\rightarrow$ ``so that the game would be known as the "Super Bowl of the Century.''
\end{itemize}

\paragraph{2) There should be at least one long edit.}

Among all your edits, there should be at least one long edit, which rewrites the whole sentence or at least half of the sentence. 

In the above example, the long edit is: ``on February 7, 2016, at Levi's Stadium in the San Francisco Bay Area at Santa Clara, California.'' $\rightarrow$ ``Mercedes-Benz Superdome in New Orleans, Louisiana and was the first Super Bowl to be played in the United States.''

\paragraph{3) The edits should create contradicting information. }

After your edits, the original passage and the modified passage should have contradicting information. One way to test it is that: when you ask questions about your modified information, the original passage and the modified passage gives contradicting answers. 

For example: after you edit ``Denver Broncos'' to ``San Francisco 49ers'', the original and modified passages are shown in the Figure below: 


\vspace{0.4cm}

Original Text: 

\texttt{The American Football Conference (AFC) champion \textbf{Denver Broncos} defeated the National Football Conference (NFC) champion Carolina Panthers 24-10 to earn their third Super Bowl title.} 

Modified Text: 

\texttt{The American Football Conference (AFC) champion \textbf{San Francisco 49ers} defeated the National Football Conference (NFC) champion Carolina Panthers 24-10 to earn their third Super Bowl title.} 

\vspace{0.4cm}

When you ask the question: ``Which NFL team won Super Bowl 50?”, the original passage gives you the answer ``Denver Broncos'', and the modified passage gives you the answer ``San Francisco 49ers''. This is a contradiction. 

Another example is the following edit: ``so that the logo could prominently feature the Arabic numerals 50.'' $\rightarrow$ ``so that the game would be known as the ``Super Bowl of the Century''.


\vspace{0.4cm}

Original Text: 

\texttt{... the league emphasized the "golden anniversary" with various gold-themed initiatives, as well as temporarily suspending the tradition of naming each Super Bowl game with Roman numerals (under which the game would have been known as "Super Bowl L"), \textbf{so that the logo could prominently feature the Arabic numerals 50. }}

Modified Text: 

\texttt{... the league emphasized the "golden anniversary" with various gold-themed initiatives, as well as temporarily suspending the tradition of naming each Super Bowl game with Roman numerals (under which the game would have been known as "Super Bowl L"), \textbf{so that the game would be known as the "Super Bowl of the Century". }}

\vspace{0.4cm}

When you ask the question: ``Why the league suspended the tradition of naming Super Bowls with Roman numerals?'' the original passage and the modified passage also give you contradicting answers. 

However, the following passage does \textbf{NOT} create any contradiction, because the modified information is just a paraphrasing of the original information. 


\vspace{0.4cm}

Original Text: 

\texttt{The American Football Conference (AFC) champion Denver Broncos defeated the National Football Conference (NFC) champion Carolina Panthers 24-10 to \textbf{earn their third Super Bowl title.} }

Modified Text: 

\texttt{The American Football Conference (AFC) champion Denver Broncos defeated the National Football Conference (NFC) champion Carolina Panthers 24-10 to \textbf{win the Super Bowl.} }

\vspace{0.4cm}

\paragraph{4) The edits should modify important information in the passage.}

Your edits should focus on important information in the passage, \textit{i.e.}, points that people are usually interested in and would usually ask about. For example, time, location, purpose, outcome, reason, etc. Please avoid editing trivial and unimportant details. 

\begin{figure*}[!t]
    \centering
    \includegraphics[width=16cm]{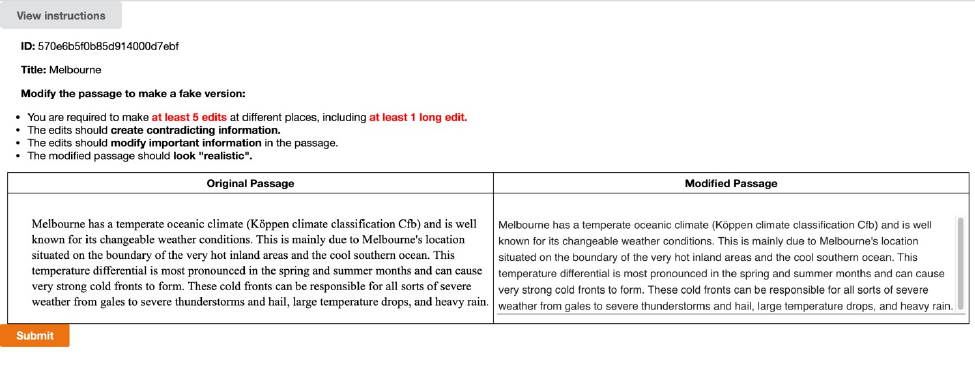}
    \caption{The annotation interface in the Amazon Mechanical Turk. }
\end{figure*}

For example, the following trivial edit is not supported: 

\vspace{0.4cm}

Original Text: 

\texttt{the game would have been known as "Super Bowl \textbf{L}"...} 

Modified Text: 

\texttt{the game would have been known as "Super Bowl \textbf{H}"...} 

\vspace{0.4cm}

\paragraph{5) The modified passage should look ``realistic''. }

The final modified passage should look ``realistic''. Don’t make obvious logic or commonsense mistakes to make the reader easily know that this is a fake passage by simply going through it. 

For example, the following edit is not supported. 

\vspace{0.4cm}

Original Text: 

\texttt{The game was played on February 7, 2016, at Levi's Stadium in the San Francisco Bay Area at \textbf{Santa Clara}, California.}

Modified Text: 

\texttt{The game was played on February 7, 2016, at Levi's Stadium in the San Francisco Bay Area at \textbf{New York City}, California.}

People can easily tell the modified passage is fake since everybody knows that New York is not a city in California. 

\vspace{0.4cm}

\subsection{Annotation Interface}

The original passage is shown on the left for your reference, you should modify the passage in the text box on the right to make the fake passage. After you finished the edits, Click ``Submit''.



\end{document}